\pdfoutput=1

\documentclass[11pt]{article}

\usepackage{naacl2021}

\usepackage{times}
\usepackage{latexsym}

\usepackage[T1]{fontenc}

\usepackage[utf8]{inputenc}

\usepackage{microtype}

\usepackage{booktabs}
\usepackage{amsmath,amssymb}
\usepackage{graphicx}
\usepackage{array}
\usepackage{enumitem}
\usepackage{color}
\usepackage{caption}
\usepackage{subcaption}
\usepackage{pifont}
\usepackage{multirow}
\newcolumntype{L}[1]{>{\raggedright\let\newline\\\arraybackslash\hspace{0pt}}m{#1}}
\newcolumntype{C}[1]{>{\centering\let\newline\\\arraybackslash\hspace{0pt}}m{#1}}
\newcolumntype{R}[1]{>{\raggedleft\let\newline\\\arraybackslash\hspace{0pt}}m{#1}}

\usepackage{listings}
\usepackage{xcolor}

\definecolor{codegreen}{rgb}{0,0.6,0}
\definecolor{codegray}{rgb}{0.5,0.5,0.5}
\definecolor{codepurple}{rgb}{0.58,0,0.82}
\definecolor{backcolour}{rgb}{0.95,0.95,0.92}

\lstdefinestyle{mystyle}{
    backgroundcolor=\color{backcolour},   
    commentstyle=\color{codegreen},
    keywordstyle=\color{magenta},
    numberstyle=\tiny\color{codegray},
    stringstyle=\color{codepurple},
    basicstyle=\ttfamily\footnotesize,
    breakatwhitespace=false,         
    breaklines=true,                 
    captionpos=b,                    
    keepspaces=true,                 
    numbers=left,                    
    numbersep=5pt,                  
    showspaces=false,                
    showstringspaces=false,
    showtabs=false,                  
    tabsize=2
}

\title{Hierarchical Transformer for Task Oriented Dialog Systems}

\author{Bishal Santra\thanks{\ \ Equal Contributions} \\
    bsantraigi\textsuperscript{$\dagger$} \\\And
  Potnuru Anusha\footnotemark[1] \\
    anusha.sparkx\textsuperscript{$\dagger$}\\\And
  Pawan Goyal \\
    pawang\textsuperscript{$\ddagger$}\\\AND
  {\normalfont Computer Science and Engineering Dept.} \\
  Indian Institute of Technology Kharagpur \\
  Kharagpur, W.B., India \\
  \texttt{\{$\dagger$\}@gmail.com, \{$\ddagger$\}@cse.iitkgp.ac.in}
}

\begin{document}
\maketitle
\begin{abstract}

Generative models for dialog systems have gained much interest because of the recent success of RNN and Transformer based models in tasks like question answering and summarization. Although the task of dialog response generation is generally seen as a sequence to sequence (Seq2Seq) problem, researchers in the past have found it challenging to train dialog systems using the standard Seq2Seq models. Therefore, to help the model learn meaningful utterance and conversation level features, \citet{sordoni-etal-2015-neural, serban2016building} proposed Hierarchical RNN architecture, which was later adopted by several other RNN based dialog systems. With the transformer-based models dominating the seq2seq problems lately, the natural question to ask is the applicability of the notion of hierarchy in transformer based dialog systems. 
In this paper, we propose a generalized framework for Hierarchical Transformer Encoders and show how a standard transformer can be morphed into any hierarchical encoder, including HRED and HIBERT like models, by using specially designed attention masks and positional encodings. 
We demonstrate that Hierarchical Encoding helps achieve better natural language understanding of the contexts in transformer-based models for task-oriented dialog systems through a wide range of experiments. The code and data for all experiments in this paper has been open-sourced\footnote{Experiments in this paper: \url{https://github.com/bsantraigi/HIER}} \footnote{PyTorch implementation of Hierarchical Transformer Encoder: \url{https://github.com/bsantraigi/hier-transformer-pytorch}}. 


\end{abstract}

\section{Introduction}

Dialog systems are concerned with replicating the human ability to make conversation. In a generative dialog system, the model aims at generating coherent and informative responses given a dialog context and, optionally, some external information through knowledge bases \cite{wen-etal-2017-network} or annotations e.g. belief states, dialog acts etc. \cite{chen-etal-2019-semantically, zhao-etal-2017-learning}. 

A dialog is usually represented as a series of utterances. However, it is not sufficient to view each utterance independently for engaging in a conversation. In a dialogue between humans, the speakers communicate both utterance level and dialog level information. E.g., dialog intent often cannot be detected by looking at a single utterance, whereas dialog acts are specific to each utterance and change throughout a conversation. Intuitively, we can instruct the model to achieve both utterance level and dialog level understanding separately through a hierarchical encoder \cite{serban2016building}. 


 
There has been a lot of interest in the past towards using the Hierarchical Encoder-Decoder (HRED) model for encoding utterances in many RNN based dialog systems. However, since the rise of Transformers and self-attention \cite{vaswani2017attention}, the use of hierarchy has not been explored further for transformer-based dialog models. Past research and user-studies have also shown that hierarchy is an important aspect of human conversation \cite{jurafsky2000speech}. But, most previous works based on transformer have focused on training models either as language models \cite{budzianowski_2019_gpt2, zhang2019dialogpt} or as standard (non-hierarchical) Seq2Seq models \cite{chen-etal-2019-semantically, zhang2019taskoriented, wang-etal-2020-multi} with certain task specific extensions. Although arguably, the self-attention mechanism might automatically learn such a scheme during the training process, our empirical results show that forcing this inductive bias by manual design as proposed here leads to better performing models. 



This paper bridges these two popular approaches of transformers and hierarchical encoding for dialogs systems to propose a family of Hierarchical Transformer Encoders. Although arguably, the self-attention mechanism of standard encoders might automatically learn such a scheme during the training process, our empirical results show that forcing this inductive bias by manual design as proposed here leads to better performing models. Our contributions in this paper include:
\begin{itemize}[noitemsep, topsep=0pt]
    \item We propose a generalized framework for hierarchical encoders in transformer based models that covers a broader range of architectures including existing encoding schemes like HRED/HIBERT \cite{zhang-etal-2019-hibert} and possibly other novel variants. We call members of this family of hierarchical transformer encoders as an \textbf{HT-Encoder}.
    \item Then, we formulate a straightforward algorithm for converting an implementation of standard transformer encoder into an HT-Encoder by changing the attention mask and the positional encoding.
    \item Building upon that, we show how an HRED/HIBERT like hierarchical encoder (\textbf{HIER-CLS}) can be implemented using our HT-Encoder framework.
    \item We also showcase a novel HT-Encoder based model, called \textbf{HIER}, with a context encoding mechanism different from HRED. We show that these simple HT-Encoder based baselines achieve at par or better performance than many recent models with more sophisticated architectures or training procedures. We make a thorough comparison with many recently proposed models in four different experimental settings for dialog response generation task.
    \item We further apply HT-Encoder to a state-of-the-art model, Marco \cite{wang-etal-2020-multi}, for task-oriented dialog systems and obtain improved results.
\end{itemize}


\begin{figure}[ht]
     \centering
     \includegraphics[width=\linewidth]{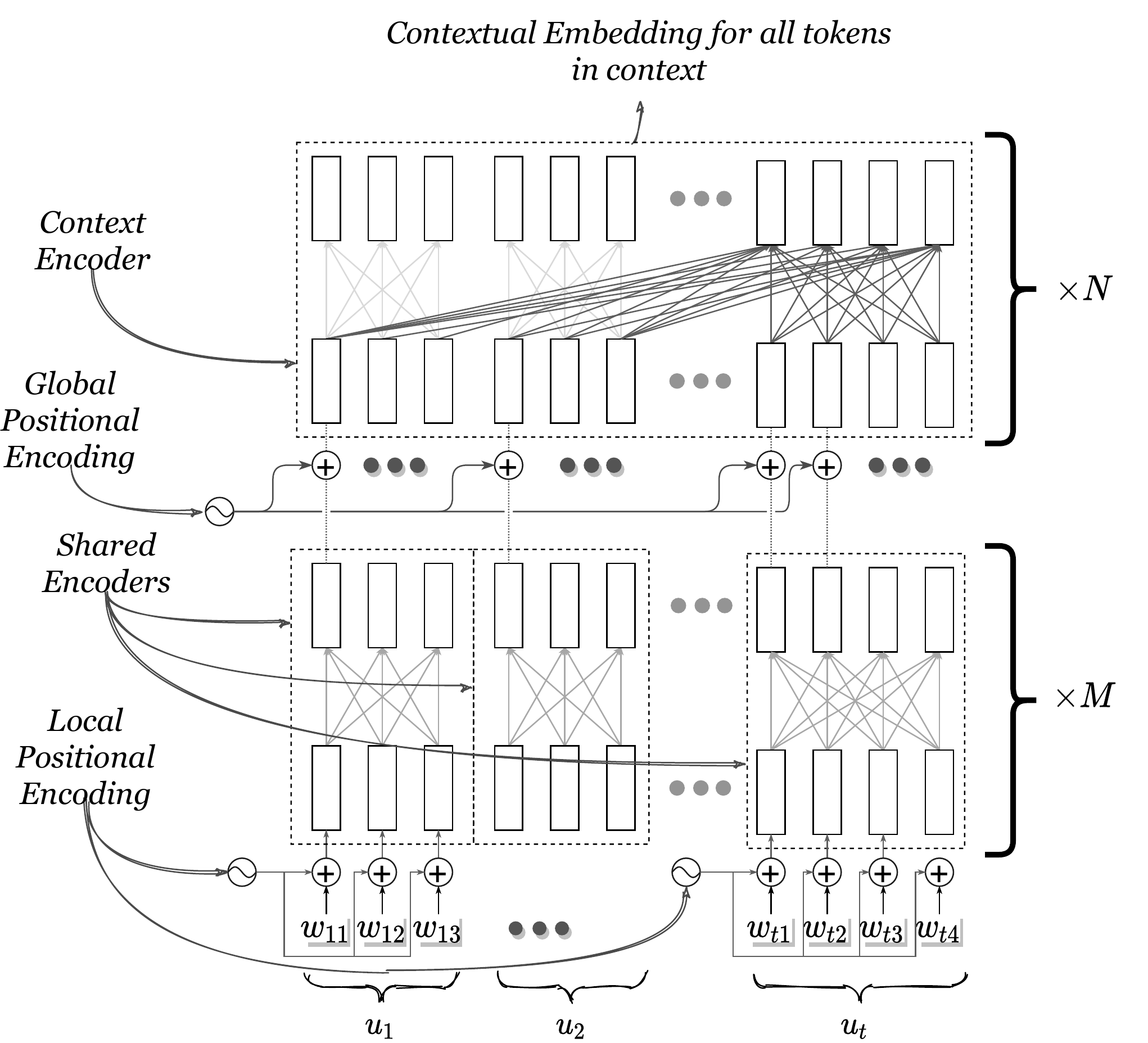}
	 \caption{Detailed architecture for a \textbf{Hierarchical Transformer Encoder} or \textbf{HT-Encoder}: The main inductive bias incorporated in this model is to encode the full dialog context hierarchically in two stages. This is done by the two encoders, 1) Shared Utterance Encoder (M layers) and 2) Context Encoder (N layers), as shown in the figure. Shared encoder first encodes each utterance ($u_1, u_2, \dots, u_t$) individually to extract the utterance level features. The same parameterized Shared Encoder is used for encoding all utterances in the context. In the second Context Encoder the full context is encoded using a single transformer encoder for extracting dialog level features. The attention mask in context encoder decides how the context encoding is done and is a choice of the user. This one depicted in the figure is for the HIER model described in Section \ref{sec:hier}. Only the final utterance in the Context Encoder gets to attend over all the previous utterances as shown. This allows the model to have access to both utterance level features and dialog level features till the last layer of the encoding process. Notation: Utterance $i$, $u_i = [w_{i1}, \dots, w_{i|u_i|}]$, $w_{ij}$ is the word embedding for $j^{th}$ word in $i^{th}$ utterance.} 
     \label{fig:hier-detail}
\end{figure}

\section{Models}
Formally, the task of a dialog system is to predict a coherent response, $r$, given a dialog context $c$. In case of a goal oriented dialog system, context $c$ might consist of dialog history, $C_t = [U_1, S_1, . . . , U_t]$, and optionally a belief state (dialog act, slot values, intent etc.) $b_t$, when available. Here, $U_i$, $S_i$ represent the user and system utterances at turn $i$, respectively. The actual target response following $C_t$ is the system utterance $S_t$. 

\subsection{Hierarchical Transformer Encoders (HT-Encoder)}
\label{sec:hte}
Like the original HRED architecture, HT-Encoder also has two basic components, a shared utterance encoder and the context encoder. Shared utterance encoder, or the \textbf{Shared Encoder} in short, is the first phase of the encoding process where each utterance is processed independently to obtain utterance level representations. In the second phase, the \textbf{Context Encoder} is used to process the full context together. These context level representations are then used for the tasks like dialog state tracking or response generation. We propose two different types of Hierarchical Encoding schemes for the transformer model. 

\paragraph{1. HIER-CLS:} When \citet{serban2016building} employed a hierarchical encoder for dialog contexts, they obtained a single representative embedding, usually the final hidden state of an RNN, for each utterance. Similarly, in HIER-CLS, the context encoder utilizes only a single utterance embedding for each utterance. We do this by taking the contextual embedding of the first token (often termed as the ``CLS'' token in transformer based models) of each utterance.

\paragraph{2. HIER:} Recent works have shown the importance of contextual word embeddings. In HIER, we consider contextual embedding of all utterance tokens as input to the context encoder. We simply concatenate the whole sequence of contextual embeddings and forward it to the context encoder.

\subsection{Conversion Algorithm: Standard Encoder to HT-Encoder}
\label{sec:conversion-hte}

\begin{figure}[h!]
    \centering
    \includegraphics[width=0.75\linewidth]{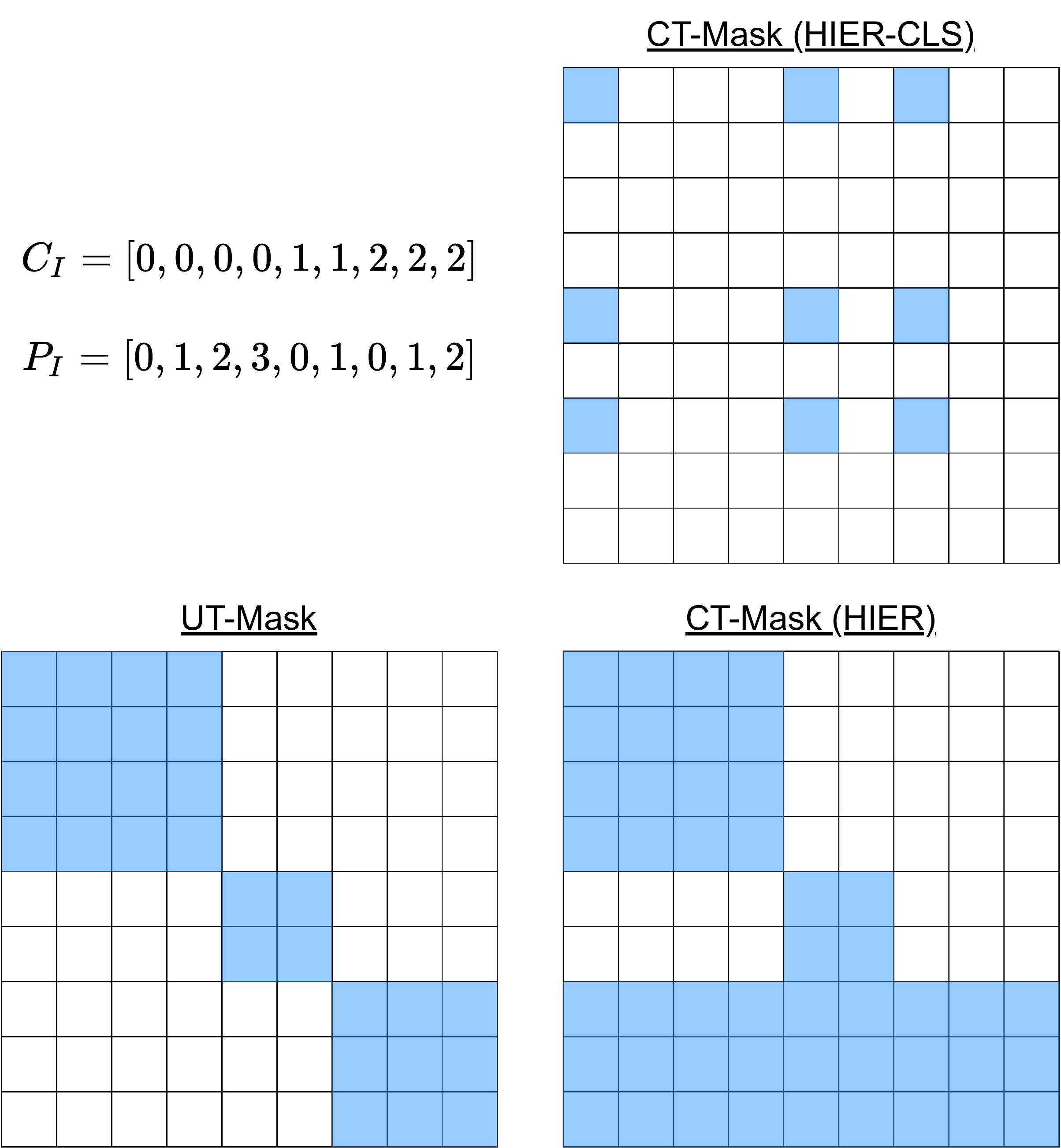}
    \caption{Example of UT-Mask ($A$ for the given $C_I$) and CT-Masks. Blue cells: 1, White cells: 0. Bottom left is the UT-Mask and on the right are CT-Masks for HIER-CLS(top) and HIER(bottom). In this example, the context comprises of three utterances of lengths $0, 1$ and $2$, respectively. $C_I$ indicates which utterance each of the tokens belongs to. The entries in $P_I$ denotes the relative position of each token with respect to utterance corresponding to it.}
    \label{fig:masks}
\end{figure}

In this section, we show how the two-step process of hierarchical encoding can be achieved using a single standard transformer encoder. If we want to have an $M$ layer utterance encoder followed by an $N$ layer context encoder, we start with an $(M+N)$ layer standard encoder. Then by applying two separate masks as designed below, we convert the standard encoder into an HT-encoder.
First, we need to encode the utterances independently. Within the self-attention mechanism of a transformer encoder, which token gets to attend to which other tokens is controlled by the attention mask. If we apply a block-diagonal mask, each block of size same as the length of utterances (as shown in Figure \ref{fig:masks} bottom-left), to the concatenated sequence of tokenized utterances, we effectively achieve the same process of utterance encoding. We call this block-diagonal mask for utterance encoding the \textbf{UT-mask}.

Similarly, another attention mask (\textbf{CT-Mask}) can explain the context encoding phase that allows tokens to attend beyond the respective utterance boundaries. See the two matrices on Figure \ref{fig:masks}'s right for examples of such CT-Masks.
From here, it can be quickly concluded that if we apply the UT-Mask for the first few layers of the encoder and the CT-Mask in the remaining few layers, we effectively have a hierarchical encoder. The CT-Mask also gives us more freedom on what kind of global attention we want to allow during context encoding. Positional encoding is applied once before utterance encoder (local PE) and once more before context encoder (global PE).

\paragraph{UT-Mask and Local Positional Encoding} The steps for obtaining the UT-Mask and positional encoding for the utterance encoder are given below and is accompanied by Figure \ref{fig:masks}. $C$ is the dialog context to be encoded. $w_{ij}$ is the $j_{th}$ token of $i_{th}$ utterance. In $C_{I}$, each index $i$ is repeated $|u_i|$ (length of $u_i$) times. And $C_{IR}$ is a square matrix created by repeating $C_I$. $P_I$ has the same dimensions as $C_I$, and it stores the position of each token $w_{ij}$ in context $C$, relative to utterance $u_i$. $\mathcal{P}:I \mapsto R^d$ is the positional encoding function that takes an index (or indices) and returns their $d$-dim positional embedding. $A$ is the UT-Mask for the given context $C$ and their utterance indices $C_I$. An example instance of this process is given in Figure \ref{fig:masks}. $\mathbf{1}(.)$ is an indicator function that returns true when the input logic holds, and is applied to a matrix or vector element-wise.
\begin{align*}
 C &= [w_{11}, w_{12}, ..., w_{Tl_T}]\\
 C_{I} &= [0, \dots, 0, 1, \dots, 1, \dots, T] \\
 P_{I} &= [0, 1, \dots, l_1 - 1, 0, \dots, l_2 - 1,\dots, l_T - 1] \\
 C_{IR} &= repeat(C_I, len(C_I), 0)\\
 A &= 1\big(2C_{IR}== (C_{IR}^T + C_{IR})\big)\\
 P_c &= \mathcal{P}[P_I, :]
\end{align*}
\paragraph{CT-Masks for Models}
The attention masks for context encoding depends on the choice for model architecture. We provide the details of the architectures and their attention masks used in our experiments in the subsequent section. There are other masks possible, but these are the ones we found to be working best in their respective settings.

\begin{figure}[t!]
     \centering
     \begin{subfigure}[b]{\linewidth}
           \centering
	    \includegraphics[width=\linewidth]{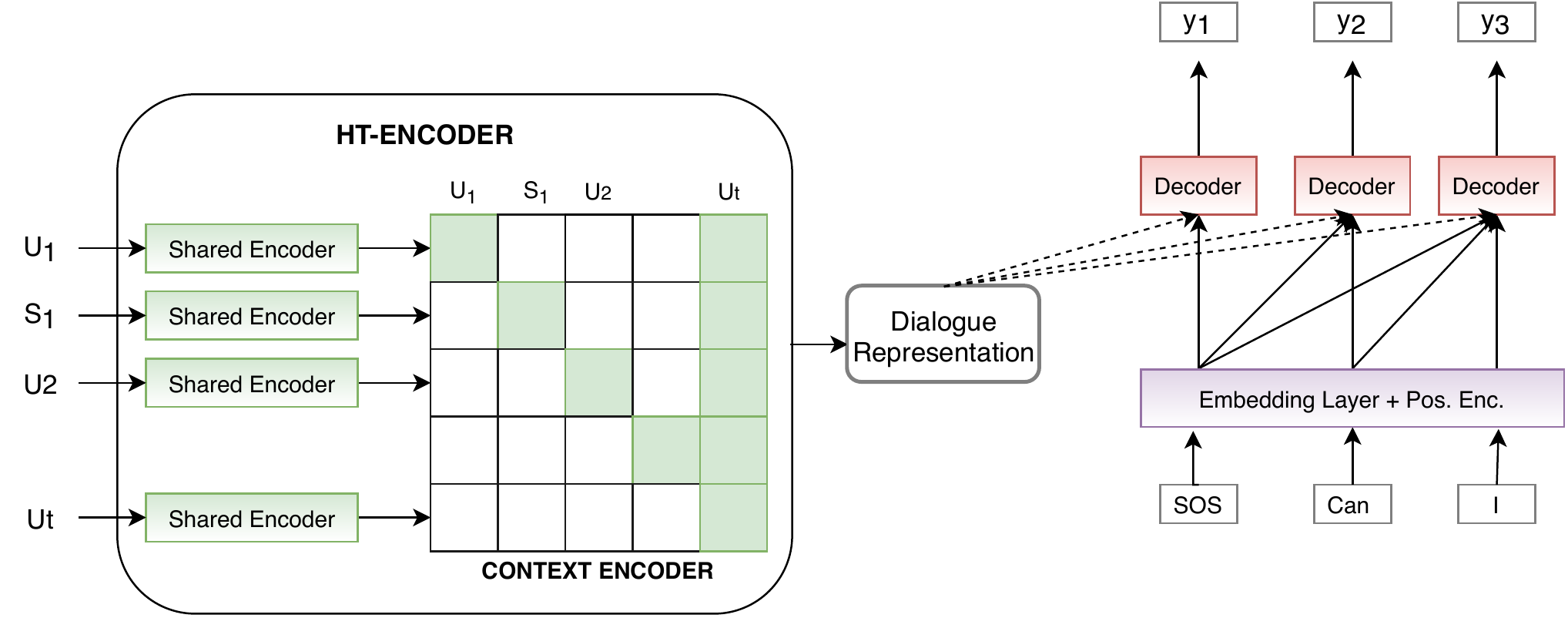}
	     \caption{\textbf{Model: HIER}}
	    \label{fig:hier}
	    \end{subfigure}
     \hfill
     \begin{subfigure}[b]{\linewidth}
         \centering
	    \includegraphics[width=\linewidth]{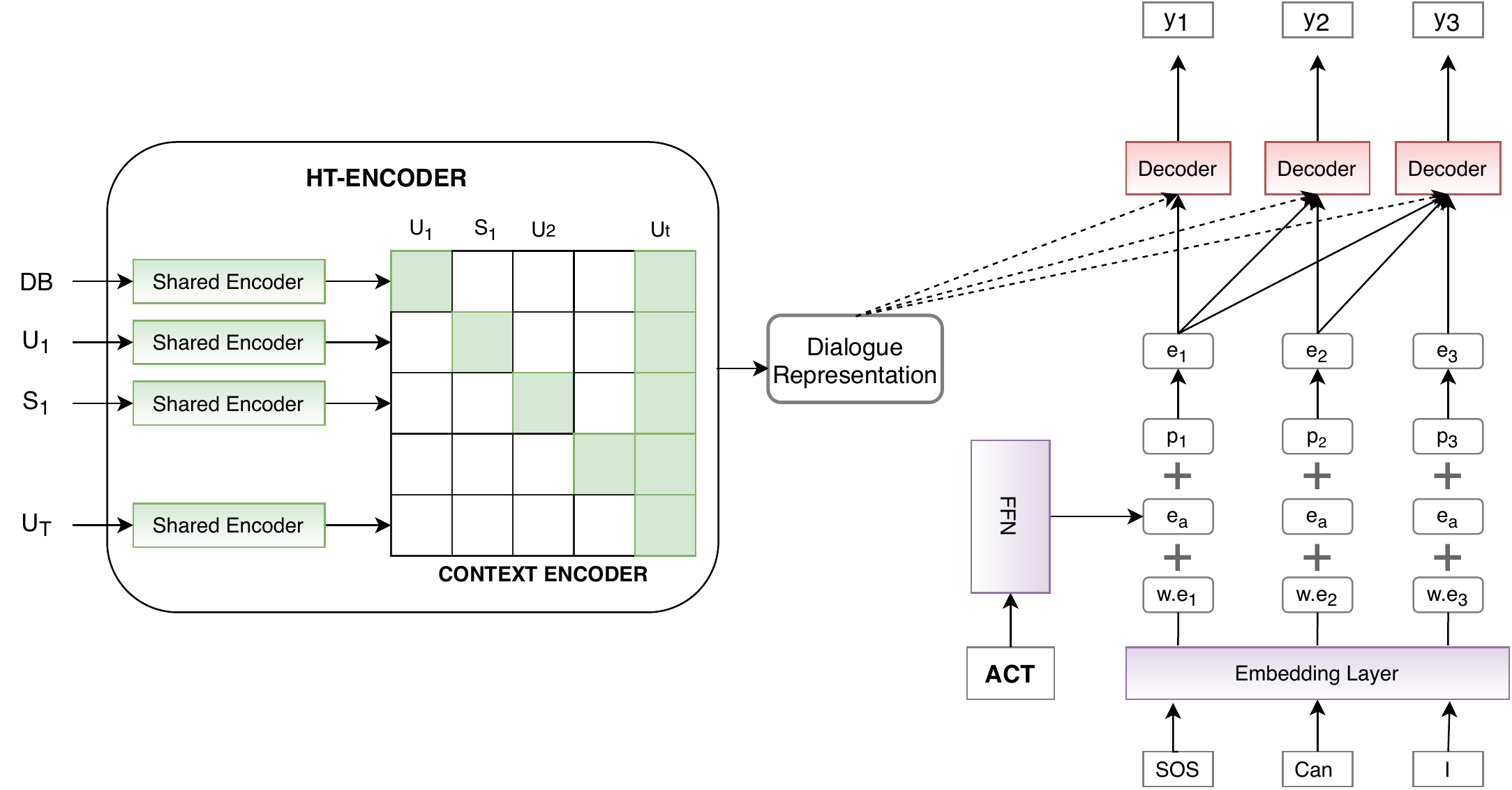}
	     \caption{\textbf{Model: HIER++}}
	    \label{fig:hier++}
     \end{subfigure}
     \caption{The proposed architecture for the hierarchical transformer: (a) HIER: when the belief states are not available and (b) HIER++: when the belief states are available.}
     \label{fig:models-part2}
\end{figure}

\subsection{Model Architectures}
\label{sec:hier}
We propose several model architectures to test the effectiveness of the proposed HIER-Encoder in various experimental settings. These architectures are designed to fit well with the four experimental settings (see Section \ref{sec:four-eval}) of the response generation task of the MultiWOZ dataset in terms of input and output.

The tested model architectures are as follows. Using the HIER encoding scheme described in Section \ref{sec:hte}, we test two model architectures for response generation, namely HIER and HIER++.

\paragraph{HIER:} HIER is the most straightforward model architecture with an HT-Encoder replacing the encoder in a Transformer Seq2Seq. The working of the model is shown in Figure \ref{fig:hier}. First, in the utterance encoding phase, each utterance is encoded independently with the help of the UT-Mask. In the second half of the encoder, we apply a CT-Mask as depicted by the figure's block attention matrix. Block $B_{ij}$ is a matrix which, if all ones, means that utterance $i$ can attend to utterance $j$'s contextual token embeddings. The local and global positional encodings are applied, as explained in Section \ref{sec:conversion-hte}. A standard transformer decoder follows the HT-Encoder for generating the response.

The CT-Mask for HIER was experimentally obtained after trying a few other variants. The intuition behind this mask was that the model should reply to the last user utterance in the context. Hence, we design the attention mask to apply cross attention between all the utterances and the last utterance (see Figure \ref{fig:hier}).

\paragraph{HIER++:}
\label{sec:abl3}
HIER++ is the extended version of the HIER model, as shown in Figure \ref{fig:hier++}, that also takes the dialog act label as input.
The dialog act representation proposed in \citet{chen-etal-2019-semantically} consists of the domain, act, and slot values. A linear feed-forward layer (FFN) acts as the embedding layer for converting their 44-dimension multi-hot dialog act representation. The output embedding is added to the input token embeddings of the decoder in HIER++ model. Similar to HDSA, we also use ground truth dialog acts during training, and predictions from a fine-tuned BERT model during validation and testing.
HIER++ is applied to the Context-to-Response generation task of the MultiWOZ dataset.  

\paragraph{HIER-CLS:} As described in Section \ref{sec:hte}, the encoding scheme of HIER-CLS is more akin to the HRED \cite{chen-etal-2019-semantically} and HIBERT \cite{zhang-etal-2019-hibert} models. It differs from HIER++ only with respect to the CT-Mask. 


\paragraph{Ablations} To understand the individual impact of UT-Mask and CT-Mask, we ran the same experiments with the following model ablations. 
\begin{enumerate}
 \item \textbf{SET:} HIER without the context encoder. Each utterance is encoded independently. It shows the importance of context encoding. Effectively, this model is only the shared utterance encoder (SET) applied to each utterance independently.
 \item \textbf{MAT:} HIER without the utterance encoder. This model only uses the context encoder as per the context attention mask of Figure \ref{fig:hier}. As this is equivalent to a simple transformer encoder with a special attention mask, we call it the Masked Attention Transformer or MAT. 
 \item \textbf{SET++:} An alternative version of SET with dialog-act input to the decoder similar to HIER++.
\end{enumerate}

\paragraph{HIER-Joint:}
Finally, we propose the HIER-Joint model\footnote{Block diagram for HIER-Joint model has been provided in supplementary material.} suitable for the end-to-end response generation task of the MultiWOZ dataset. The HIER-Joint model comprises an HT-Encoder and three transformer decoders for decoding belief state sequence, dialog act sequence, and response. It is jointly trained to predict all three sequences simultaneously. As belief state labels can help dialog-act generation, and similarly, both belief and act labels can assist response generation, we pass the token embedding from the belief decoder and act decoder to the response decoder. Act decoder receives mean token embedding from the belief decoder too.

%
%
%
%

\begin{table}[ht]
    \centering
    \begin{tabular}{ c c c c c }
    \hline
        Model & L & H & A & E  \\
        \hline
        SET & 6/-/3 & 100 & 4 & 100 \\
        MAT & -/4/6 & 200 & 5 & 100 \\
        HIER & 3/3/3 & 100 & 4 & 100 \\
        SET++ & 4/-/3 & 91 & 7 & 175 \\
        HIER++ & 4/6/3 & 91 & 7 & 175 \\
    \hline
    \end{tabular}
    \caption{Best Hyper-parameters: L: a/b/c = number of layers in shared encoder/ Context Encoder / decoder,  H = hidden size, A = attention heads, E = embedding size.}
    \label{tab:parameters}
\end{table}

\section{Experimenal Framework} 


Our implementation is based on the PyTorch library. 
All the models use a vocabulary of size 1,505. We generate responses using beam search\footnote{\url{https://github.com/OpenNMT/OpenNMT-py/tree/master/onmt/translate}} with beam width 5. The model optimizes a cross entropy loss. Full details of model parameters are given in suplementary material.

\paragraph{Dataset} We use MultiWOZ\footnote{MultiWOZ v2.0 \url{https://github.com/budzianowski/multiwoz/blob/master/data/MultiWOZ_2.0.zip}}~\cite{budzianowski-etal-2018-multiwoz}, a multi-domain task-oriented dataset. It contains a total of 10,400 English dialogs divided into training (8,400), validation (1,000) and test (1,000). Each turn in the dialog is considered as a prediction problem with all utterances upto that turn as the context.\footnote{\label{note1}See supplementary for more details.} 

\paragraph{Baselines} To fully grasp the effectiveness of our proposed approaches, we consider several baseline\footnotemark[3] models with varying complexity and architectures. Token-MoE \cite{pei2019tokenmoe} is a token level mixture-of-experts (MoE) model. It builds upon the base architecture of LSTM-Seq2Seq with soft attention. In the decoding phase, they employ $k$ expert decoders and a chair decoder network which combines the outputs from the experts. Attn-LSTM \cite{budzianowski-etal-2018-multiwoz} uses an LSTM Seq2Seq model with attention on encoded context utterance, oracle belief state and DB search results. HRED \cite{serban2017hierarchical} model is based on the same idea of hierarchical encoding in RNN Seq2Seq networks (results source: \citealp{peng2019teacher, peng2020mtss}). The transformer based baseline \cite{vaswani2017attention} concatenates the utterances in dialog context to obtain a single source sequence and treats the task as a sequence transduction problem. HDSA \cite{chen-etal-2019-semantically} uses a dialog act graph to control the state of the attention heads of a Seq2Seq transformer model. \citet{zhang2019taskoriented} proposes to augment the training dataset by building up a one-to-many state-to-action map, so that the system can learn a more balanced distribution for the action prediction task. Using this method they train a domain-aware multi-decoder (DAMD) network for predicting belief, action and response, jointly. As each agent response may cover multiple domains, acts or slots at the same time, Marco \cite{wang-etal-2020-multi} learns to generate the response by attending over the predicted dialog act sequence at every step of decoding. SimpleTOD \cite{hosseini2020simple} and SOLOIST \cite{peng2020soloist} are both based on the GPT-2 \cite{radford2019language} architecture. The main difference between these two architectures is that SOLOIST further pretrains the GPT-2 model on two more dialog corpus before fine-tuning on MultiWOZ dataset.

\subsection{Task Settings:}
\label{sec:four-eval}
Following the literature \cite{zhang2019taskoriented, peng2020soloist}, we now consider four different settings for evaluating the strength of hierarchical encoding. 

\paragraph{1. No Annotations} First, to simply gauge the benefit of using a Hierarchical encoder in a Transformer Seq2Seq model, we compare the performance of HIER to other baselines including HRED and vanilla Transformer without any belief states and dialog act annotations.

\paragraph{2. Oracle Policy} In this setting, several recently proposed model architectures for the response generation task of MultiWOZ are compared against each other in presence of ground truth belief state and dialog act annotations. This experiment helps us understand the models' capabilities towards generating good responses (BLEU score) when true belief state and(or) dialog acts are available to them.

\paragraph{3. Context-to-Response} The model is given true belief states and DB search results in this experiment, but they need to generate the dialog act and response during inference. Some of the baselines generate dialog act as an intermediate step in their architecture whereas others use a fine-tuned BERT model.

\paragraph{4. End-to-End} This is the most realistic evaluation scheme where a model has to predict both belief states and dialog act (or one of these as per the models input requirement) for searching DB or generating response.

\subsection{Evaluation Metrics}
We used the official evaluation metrics\footnote{\url{https://github.com/budzianowski/multiwoz}} released by the authors of the MultiWOZ dataset \citep{budzianowski-etal-2018-multiwoz}: \textbf{Delexicalized-BLUE score}, \textbf{INFORM rate} (measures how often the entities provided by the system are correct),  \textbf{SUCCESS rate} (reflects how often the system is able to answer all the requested attributes), \textbf{Entity-F1 score}~\cite{wen-etal-2017-network} (measures the entity coverage accuracy), and \textbf{Combined Score} ($S=BLEU+0.5\times(Inform + Success)$) to measure the overall quality.

\paragraph{Training} Cross-entropy losses over the ground truth response and/or belief and act sequences are used for the training the models. We did hyperparameter search using the Optuna library \citep{optuna_2019} by training the model upto 5 epochs. Final models were trained
\footnote{A system with two Tesla P100 GPUs were used for training.} upto 30 epochs with early stopping. 


\section{Results}
For the four different experimental settings discussed in Section \ref{sec:four-eval}, we showcase results from those experiments in Tables \ref{tab:overall_results} through \ref{tab:overall_results4}. Table \ref{tab:overall_results} shows the results from our experiments when no oracle is present. By comparing the performance of Transformer, SET and MAT baselines against that of HIER we can see that in each case HIER is able to improve in terms of BLEU, Success and overall Score. HIER being better than SET and MAT implies that only the UT-Mask or the CT-Mask is not sufficient, the full scheme of HT-Encoder is necessary for the improvement. 
The exception in the improvements is the SET model which has the highest inform score of 76.80. Although, we observe that it is the combination of the BLEU and Inform score that depicts the real quality of the responses. As BLEU measures precision of n-grams and inform measures recall of task related entities, only when both metrics increase we get a better performing model. This is reflected \textit{upto some extent} in Entity-F1 score (H-Mean of entity recall and precision), but it too ignores tokens other than task related entities. So SET only having a higher inform score may mean that it is over-predicting some entities leading to improved recall.

\begin{table*}[h!]
\centering
    \begin{tabular}{lccccc}
    \toprule
    \multirow{2}{*}{Models} & \multicolumn{5}{c}{Evaluation Metrics}\\
    \cmidrule(l){2-6}
    &BLEU& Entity-F1 & Inform & Success & Score \\ \midrule
    HRED & 17.50 & - & 70.7 & 60.9 & 83.3  \\
    TokenMoE  & 16.81 & - & 75.30 & 59.70 & 84.31 \\
    Transformer  & 19.1 & 55.1 & 71.1 & 59.9 & 84.60  \\
    \textbf{SET}  & 18.67 & 51.61 & 76.80 & 57.69 & 85.92  \\
    \textbf{MAT}  & 18.86 &	54.89 &	71.9 &	52.5 &	81.06  \\
    \textbf{HIER}  & 20.91 & 54.45 & 73.60 & 60.10 & 87.76  \\
    \bottomrule
    \end{tabular}
    \caption{Simplest Baselines in absence of both Belief or Policy / Dialog Act annotations}
    \label{tab:overall_results}
\end{table*}

\begin{table*}[ht]
\centering
    \begingroup
    \setlength{\tabcolsep}{4.5pt} 
        \begin{tabular}{lccccccccc}
        \toprule
        \multirow{2}{*}{Models} & \multirow{2}{*}{Pretraining} & \multicolumn{3}{c}{Annotations} & \multicolumn{5}{c}{Evaluation Metrics}\\
        \cmidrule(l){3-5} \cmidrule(l){6-10}
        &&Belief&DB&Policy&BLEU& Entity-F1 & Inform & Success & Score \\ \midrule
        SimpleTOD & GPT-2 & Oracle & Oracle & Oracle & 17.78 & - & 93.4 & 83.2 & 106.08 \\
        SimpleTOD &  GPT-2 & Oracle & - & Oracle & 18.61 & - & 92.3 & 85.8 & 107.66 \\
        HDSA & - & Oracle & Oracle & Oracle & 30.4 & 86.2 & 87.9 & 78.0 & 113.4 \\
        DAMD & - & Oracle & Oracle & Oracle & 27.3 & - & 95.4 & 87.2 & 118.5 \\
        \textbf{SET++} & - & - & - & Oracle & 25.56 &	82.27 &	85.7 &	74.3 &	105.56  \\
        \textbf{HIER++} & - & - & - & Oracle & 29.54 & 85.01 & 88.3 & 85.4 & 116.39  \\
        \textbf{HIER-CLS} & - & - & - & Oracle & 29.29 & 84.23 & 88.3 & 85.9 & 116.39 \\
        \bottomrule
        \end{tabular}
    \endgroup
    \caption{Context-to-Response generation with Oracle Policy. Superior Performance of DAMD: DAMD always receives an extra input of $B_{t-1}$ annotation, while predicting for $B_t$ or response $R_t$, which helps in NLU of the subsequent utterances. This is not available in any other models.}
    \label{tab:overall_results2}
\end{table*}

\begin{table*}[ht!]
\centering
    \begingroup
    \setlength{\tabcolsep}{4.5pt} 
        \begin{tabular}{lccccccccc}
        \toprule
        \multirow{2}{*}{Models} & \multirow{2}{*}{Pretraining} & \multicolumn{3}{c}{Annotations} & \multicolumn{5}{c}{Evaluation Metrics}\\
        \cmidrule(l){3-5} \cmidrule(l){6-10}
        &&Belief&DB&Policy&BLEU& Entity-F1 & Inform & Success & Score \\ \midrule
        AttLSTM & - & Oracle & Oracle & - & 18.80 & 54.8 & 71.2 & 60.2 & 84.50  \\
        SimpleTOD & GPT-2 & Oracle & Oracle & Gen & 16.9 & - & 84 & 72.8 & 94.5 \\
        HDSA & - & Oracle & Oracle & BERT & 23.6 & 68.9 & 82.9 & 68.9 & 99.50 \\
        DAMD & - & Oracle & Oracle & Gen &	18.60 & - &	89.20 & 77.90 & 102.15 \\
        SOLOIST & GPT-2, DC & Oracle & Oracle & - & 18.03 & - & 89.60 & 79.30 & 102.49 \\
        Marco & - & Oracle & Oracle & Gen & 19.45 & - & 90.30 & 75.20 & 102.20 \\
        Marco-BERT & - & Oracle & Oracle & BERT & 20.02	& 59.99 & 92.3 & 78.6	& 105.47 \\
        \textbf{SET++} & - & Oracle & Oracle & BERT & 22.08 & 65.33 & 86.2 & 76.3 & 103.33 \\
        \textbf{HIER++} & - & Oracle & Oracle & BERT & 23.04 & 64.15 & 86.5 & 76.6 & 104.59 \\
        \textbf{HIER-CLS} & - & Oracle & Oracle & BERT & 22.89 & 64.57 & 85.2 & 76.8 & 103.89 \\
        \bottomrule
        \end{tabular}
    \endgroup
    \caption{Context-to-Response: For this experiment only belief-states are given. GPT-2,DC means a second pretraining phase using extra dialog corpus (DC) starting from GPT-2 model parameters.}
    \label{tab:overall_results3}
\end{table*}

\begin{table*}[ht!]
\centering
    \begingroup
    \setlength{\tabcolsep}{4.5pt} 
        \begin{tabular}{lccccccccc}
        \toprule
        \multirow{2}{*}{Models} & \multirow{2}{*}{Pretraining} & \multicolumn{3}{c}{Annotations} & \multicolumn{5}{c}{Evaluation Metrics}\\
        \cmidrule(l){3-5} \cmidrule(l){6-10}
        &&Belief&DB&Policy&BLEU& Entity-F1 & Inform & Success & Score \\ \midrule
        DAMD & - & Gen* & Oracle & Gen & 16.60 & - & 76.40 & 60.40 & 85.00 \\
        SimpleTOD & GPT-2 & Gen & - & Gen & 15.01 & - & 84.4 & 70.1 & 92.26 \\
        SOLOIST & GPT-2, DC & Gen & Gen & - & 16.54 & - & 85.50 & 72.90 & 95.74 \\
        \textbf{HIER-Joint} & - & Gen & - & Gen  & 19.74 & 53.94 & 80.5 & 71.7 & 95.84 \\
        \bottomrule
        \end{tabular}
    \endgroup
    \caption{End-to-End: Belief State predicted by model itself. *In the End-to-End setting also, DAMD will need to use the oracle $B_{t-1}$ for predicting the current belief $B_t$.}
    \label{tab:overall_results4}
\end{table*}

\begin{table*}[h!] 
\centering
    \begingroup
    \setlength{\tabcolsep}{3.5pt} 
        \begin{tabular}{lccccccc}
        \toprule
        \multirow{2}{*}{Models} & \multicolumn{3}{c}{Act Prediction} & \multicolumn{4}{c}{Response Generation}\\
        \cmidrule(l){2-4} \cmidrule(l){5-8}
        & Precision & Recall & F1 & BLEU & Inform & Success & Score \\ \midrule
        Marco & 72.61 & 74.98 & 73.72 & 19.16 & 88.45 & 73.5 & 100.14 \\
        Marco + HTEncoder & 73.23 & 74.11 & 73.68 & 19.05 & \textbf{91.72} & \textbf{75.8} & \textbf{102.81} \\
        \midrule
        Marco-BERT & - & - & - & 19.82 & 90.86 & 76.66 & 103.58 \\
        Marco-BERT + HTEncoder & - & - & - & 19.53 & 90.99 & \textbf{78.41} & \textbf{104.23} \\
        \bottomrule
        \end{tabular}
    \endgroup
    \caption{Comparison between vanilla Marco model and Marco + HT-Encoder with proposed HT-Encoder. Boldfaced results denote statistically significant improvement with $p < 0.05$. We didn't observe any significant improvement in act-prediction F1-Score or BLEU scores for response generation. The numbers given in the table are means of 10 different runs of each algorithm.}
    \label{tab:marco-vs-hier}
\end{table*}
In the Context-to-Response generation task with oracle policy (Table \ref{tab:overall_results2}), our HIER++ and HIER-CLS models show very strong performance and beat the HDSA model (in terms of Inform and Success rates) and even the GPT-2 based baseline SimpleTOD (in terms of BLEU and Success rate). This shows that without the intricacies of the baselines, just by applying a hierarchical encoder based model we are able to perform almost at the level of the state-of-the-art model. Compared to HIER, SimpleTOD utilizes GPT-2's pretraining, and DAMD uses attention over previous belief states and action sequences. Whereas, HIER's access to oracle policy is only through the average embedding of its tokens.

Further in Table \ref{tab:overall_results4}, we compare end-to-end generation performance of HIER-Joint with baseline models that can perform belief-state and/or dialog act generation. In terms of BLEU and combined score HIER-Joint is able to perform better than the baselines. With respect to inform and success the model outperforms the DAMD baseline.

While the above experiments focus on proving the base performance of the proposed response generation models (HIER, HIER++, HIER-CLS, and ablations), HT-Encoder can be applied to any model that uses a standard transformer encoder. Hence, in a final experiment (Table \ref{tab:marco-vs-hier}), we integrate HT-Encoder with an existing state-of-the-art model Marco. We replace the standard transformer in Marco with an HT-Encoder and rerun the context-to-response generation experiment. Introducing HT-Encoder into Marco helps improve in terms of inform (minor), success and the combined score metric. The results of this experiment show that HT-Encoder is suitable for any model architecture. 

Overall, our experiments show how useful the proposed HT-Encoder module can be for dialog systems built upon transformer encoder-decoder architecture. It is also applicable to tasks where the input sequence can be split into an abstract set of subunits (e.g., search history in Sordoni's application). We believe that our proposed approach for hierarchical encoding in transformers and the algorithm for converting the standard transformer encoder makes it an invaluable but accessible resource for future researchers working on dialog systems or similar problem statements with transformer-based architectures.

\section{Related Works}
\paragraph{Task Oriented Dialog Systems} 
Researchers identify four different subtasks for any task-oriented dialog system \cite{wen-etal-2017-network}, natural language understanding (NLU), dialog state tracking (DST), dialog act or policy generation, and Natural Language Generation (NLG). Before the advent of large scale Seq2Seq models, researchers focused on building feature-rich models with rule-based pipelines for both natural language understanding and generation. It usually required separate utterance-level and dialog-level NLU feature extraction modules. These NLU features decide the next dialog act that the system should follow. This act is then converted into a natural language response using the NLG module.
\newcite{young2013pomdp} modeled this problem as a Markov Decision Process whose state comprised of various utterance and dialog features detected by an NLU module. However, such models had the usual drawback of any pipelined approaches, error propagation. \newcite{wen-etal-2017-network} proposed using neural networks for extracting features like intent, belief states, etc. and training the NLU and NLG modules end-to-end using a single loss function. Marco \cite{wang-etal-2020-multi} and HDSA \cite{chen-etal-2019-semantically} used a finetuned BERT model as their act predictor as it often triumphs other ways to train the dialog policy network (even joint learning). HDSA is a transformer Seq2Seq model with act-controllable self-attention heads (in the decoder) to disentangle the individual tasks and domains within the network. Marco uses a soft-attention over the act sequence during the response generation process.


\paragraph{Hierarchical Encoders}
The concept of Hierarchical Encoders have been used in many different context in the past. It has been most well known in the area of dialog response generation as the HRED model. Many open domain dialog systems have used the hierarchical recurrent encoding scheme of HRED for various tasks and architectures. Hierarchical Encoder was first proposed by \cite{sordoni2015hierarchical} for using in a query suggestion system. They used it encode the user history comprising multiple queries using an Hierarchical LSTM network. \citet{serban2016building} extended this work to open domain dialog generation problems and proposed the HRED network. HRED captures the high level features of the conversation in a context RNN. Several models have adopted this approach later on, e.g. VHRED \cite{serban2017hierarchical}, CVAE \cite{zhao-etal-2017-learning}, DialogWAE \cite{gu-etal-2018-multimodal}, etc.
Another area in which researchers have proposed the use of hierarchical encoder is for processing of paragraph or long documents. \citet{li2015hierarchical} used a hierarchical LSTM network for training an autoencoder that can encode and decode long paragraphs and documents. \newcite{zhang-etal-2019-hibert} proposed HIBERT where they introduced hierarchy into the BERT architecture to remove the limitation on length of input sequence. HIBERT samples a single vector for each sentence or document segment (usually contextual embedding of CLS or EOS token) from the sentence encoder to be passed onto the higher level transformer encoder. \citet{liu2019hierarchical} applies a similar approach for encoding documents in a multi-document summarization task. 
\section{Conclusion}
This paper explored the use of hierarchy in transformer-based models for task-oriented dialog system. We started by proposing a generalized framework for Hierarchical Transformer Encoders (HT-Encoders). Using that, we implemented two models, one new model called HIER, and another HIER-CLS model by adapting the existing HIBERT architecture into our framework. We thoroughly experimented with these models in four different response generation tasks of the MultiWOZ dataset. We compared the proposed models with an exhaustive set of recent state-of-the-art models to thoroughly analyze the effectiveness of HT-Encoders. We empirically show that the basic transformer seq2seq architecture, when equipped with an HT-Encoder, outperforms many of the state-of-the-art models in each experiment. We further prove its usefulness by applying it to an existing model Marco. This work opens up a new direction on hierarchical transformers in dialogue systems where complex dependencies exist between the utterances. It would also be beneficial to explore the effectiveness of the proposed HT-Encoder when applied for various other tasks. 

\bibliography{anthology,emnlp2020}
\bibliographystyle{acl_natbib}

\end{document}